\documentclass[runningheads]{llncs}
\usepackage[T1]{fontenc}
\usepackage{graphicx}
\usepackage{hyperref}

\def\BibTeX{{\rm B\kern-.05em{\sc i\kern-.025em b}\kern-.08em
    T\kern-.1667em\lower.7ex\hbox{E}\kern-.125emX}}
\begin{document}
\title{EMCAR:
Embodied Controller for Animating Robots
}

%
%
\author{Carlos Gomez Cubero\inst{1,3}\and
Elizabeth Jochum\inst{2}\orcidID{0000-0003-2570-1906}}
\authorrunning{C. Gomez Cubero and E. Jochum}
%
\institute{CREATE, Aalborg University, Aalborg, Denmark 
\and
Department of Communication and Psychology, Aalborg University, Aalborg, Denmark 
\email{jochum@ikp.aau.dk} \url{https://www.communication.aau.dk} 
\and CyPhy Life, Robotics and AI Lab,  School of Science \& Technology, IE University, Madrid, Spain \email{carlos.gomez@ie.edu} \url{https://cyphy.life/}}
\maketitle              
\begin{abstract}
This chapter describes EMCAR, a novel software tool for programming robot motion that leverages the unique affordances of artistic practices such as puppetry and drawing to conceive, design, and program novel interactions and realize new use cases for HRI. The advantage of this no-code platform is that it expands creative applications for collaborative robots - putting robots directly in the hands of artists - and provides an inclusive environment that enables individuals with little or no technical backgrounds to engage meaningfully in collaborations and robotics research.
\end{abstract}
\section{Introduction}
As AI-enabled robots move into the real world and find new applications and use cases, there is growing interest in exploring the potential of the visual and performing arts and bringing artists into the design and development of robots and Human-Robot Interaction (HRI) research through interdisciplinary collaboration. These collaborations help to reimagine the role of robots beyond mere functional tasks.   This chapter presents an exploratory, arts-led research effort to investigate tools and practices that can facilitate meaningful collaboration between artists, computer scientists, and engineers, and individuals beyond university classrooms.  Over several years and in collaboration with research labs, ballet schools, and community centers from Australia to Denmark, we have co-developed custom hardware and software tools that allow for creative and imaginative interactions between people and robots. \textbf{The Embodied Controller for Animating Robots}
(EMCAR) is the result of a multi-year effort initiated by researchers at the RELATE Research Lab for Art and Technology at Aalborg University.  EMCAR is a custom software tool for controlling a collaborative robot arm that offers direct, embodied interaction for generating and programming animation sequences by physically manipulating the robot. This technique gives people with little technical knowledge the opportunity to work directly and intuitively with the robot as they would with other materials. The platform was inspired by a collaborative drawing program first developed by members of the research team and described in \cite{10.1007/978-3-030-62056-1_7}.

EMCAR makes generating robot performances easier by allowing people to directly puppeteer the robot, and also allowing for real-time tele-operation by using different inputs such as a digital drawing tablet where the robot's movements are remotely operated using a stylus (Figure 1). The EMCAR implementation is built on top of open-source software and available at \url{https://github.com/Carlos-biru/EMCAR-CODRA-CV}. EMCAR was subsequently stress tested in the development of a human-robot dance performance described in \cite{alma9921065050005762}, where a dancer interacts with a robot in two modes: firstly as a puppet with pre-recorded movements and secondly using  a depth camera and computer vision software that maps the dancer’s body position into the robot task space. Although it was developed as a tool for performance, EMCAR has potential for a range of diverse applications and use cases. 

\section{Who can benefit from EMCAR?}
Recently, there has been growing interest in Robots \& Art collaborations, as robotics increasingly enters domains traditionally dominated by people, such as visual art, performance, education, and therapy. Engineers and computer scientists expand their horizons by working alongside artists in interdisciplinary research teams \cite{alma9921191183905762}.
Too often, however, artists are invited into research labs merely as creative instigators. One of the most common barriers to meaningful participation is technical literacy. Without a background in robotics or programming, artists are limited in how fully they can  participate as members of the research team, and too often their contributions are often limited to suggesting their ideas or follow the development from the sidelines. The advent of tools like visual programming languages and vibe coding are making computation-based art more accessible, but robots remains out of reach for many, as the application programming interfaces (APIs) for robots require specialised knowledge, and the interfaces are not intuitive to the general population. Furthermore, when working with commercial robots, the kinds of activities and functions available are usually predefined. The motivation behind EMCAR was to bridge this gap and expand possibilities for creative exploration and collaboration by developing a low-code/no code tool to enable artists and other non-technical users to work intuitively with robots, designing and controlling robot motions and behavior through embodied interaction rather than code. 

EMCAR is a perfect match for:

\subsection{Visual and Performing Artists}
Visual artists can use EMCAR to explore new modes of art-making, such as robotic drawing or kinetic sculptures, by controlling a robot arm through natural gestures. EMCAR supports drawing and painting with different tools, allowing artists to experiment with line quality, motion, and collaboration with the robot as a co-creator. Performing artists, such as puppeteers, dancers, and theater makers can use EMCAR to choreograph movements and integrate robotics into live performances.

\subsection{Students and Educators}
In the classroom, EMCAR has proven especially effective in introducing students to robotics through creative experimentation rather than technical formalism. Core robotics concepts such as degrees of freedom, inverse kinematics, coordinate systems, and transformation frames can easily be demonstrated live, only by manipulating the robot arm. The open-ended nature of EMCAR makes it particulary well-suited for educational workshops aimed at students from non-technical backgrounds. The interface is easy to use and does not require prior experience with computer science or robotics. It encourages problem-solving, embodied learning, and motion-based storytelling, while making robotics accessible and engaging through hands-on interaction.   

\subsection{HRI and Creative Robotics Researchers}
For researchers in HRI, EMCAR offers a flexible platform for prototyping interaction modalities, collecting user data, and simulating use cases of the human-robot interaction in close proximity. Its emobodied-based approach and its Graphical User Interface (GUI) allows the researcher to quickly generate repeatable robot behaviors without extensive low-level programming. This makes EMCAR a valuable tool for Wizard of Oz studies. Moreover, because the platform is open-source and built on accessible hardware and software framework, it invites modification and extension. Researchers can adapt EMCAR for new hardware, integrate additional sensors (such as depth cameras or biosignals), or customize the interaction pipeline for specific experimental needs.

\subsection*{}
Whether in a museum, classroom, research lab, or performance hall, EMCAR reduces the barrier to entry and invites new audiences to engage with robotic systems in creative and intuitive ways.

\section{What can you do with EMCAR?}
EMCAR is intended for use by non-technical individuals making it accessible for a wider audience. The platform was co-developed with an artist-in-residence who is both a trained dancer and classical painter and illustrator, who provided valuable insights and offered different viewpoints that helped to iterate and refine the idea until we landed on a tool that could be used without any  programming experience \cite{alma9921065050005762}. EMCAR was repeatedly stress-tested with different groups of users - from university students in a humanities course, to young ballet dancers, to members of the general public at a community center.  

EMCAR integrates two primary functionalities:
\begin{itemize}
\item \textbf{Robot drawing:} Enables users to tele-operate a robot to create drawings on a canvas.
\item \textbf{Robot animation:} Facilitates the recording of dynamic movements to produce animations
\end{itemize}

Both functionalities adhere to the principles of recording, saving, and replaying, ensuring a user-friendly experience while harnessing the capabilities of robotic technology. Helped by a graphical user interface, EMCAR facilitates a seamless and intuitive operation of the robot, and allows a natural workflow that promotes creativity.

Rapid prototyping is another key advantage of using EMCAR. The ability to quickly iterate on the design of the interactions allows users to experiment with different concepts without the constraints typically associated with robotic programming. For example, an artist can sketch a drawing and immediately after record an animation of the robot nodding, replay the animations and evaluate the feeling, comment on it with colleagues, and record an alternative version in a matter of seconds. This efficiency not only enhances the creative process but also supports collaboration, as ideas can be shared and developed rapidly.

For users with programming experience or those seeking to customize their experience, EMCAR is open-source, allowing for easy editing and improvement of the code. We encourage a community-driven approach to development, where adventurous users can contribute enhancements and tailor the software to various needs.
For example, for the performance IF/THEN \cite{ifthen_2021}, the drawing functionality, that tele-operates the robot, was edited to have a different XY input, the position of a dancer tracked by a computer vision system. Rather than control the movement of the robot with a pen or paintbrush, the dancer uses their entire body to control the motion of the robot.

\section{Using EMCAR}
EMCAR is simple to use, the user first download the software from the repository \url{https://github.com/Carlos-biru/EMCAR-CODRA-CV} into a computer in the same network as the robots. After following the installation steps from the repository, the system is ready to record the robot movements by puppeteering animations or tele-operating drawings with the tablet. Finally, these movements can be replayed using its GUI.

EMCAR was originally designed to run on a collaborative robot arm (Universal Robot UR3), but is compatible with UR5 and UR10.
EMCAR harnesses the power of UR Real-Time Data Exchange protocol (RTDE), a low-latency communication interface that provides real-time control of the robot and enables tele-operation. RTDE works as a continuous loop that updates the robot's pose to match a target position in less than a second. This loop frame rate is noticeable when drawing. The pen position updates fast, but can introduce limitations when attempting very fast or highly precise drawings. EMCAR records and saves the drawings and animations at a matching frame rate on a file, which later enables real-time replay.
However, technicalities aside, the following aspects should be taken into account when using EMCAR.

\begin{figure}[tb]
\centerline{\includegraphics[width=\columnwidth]{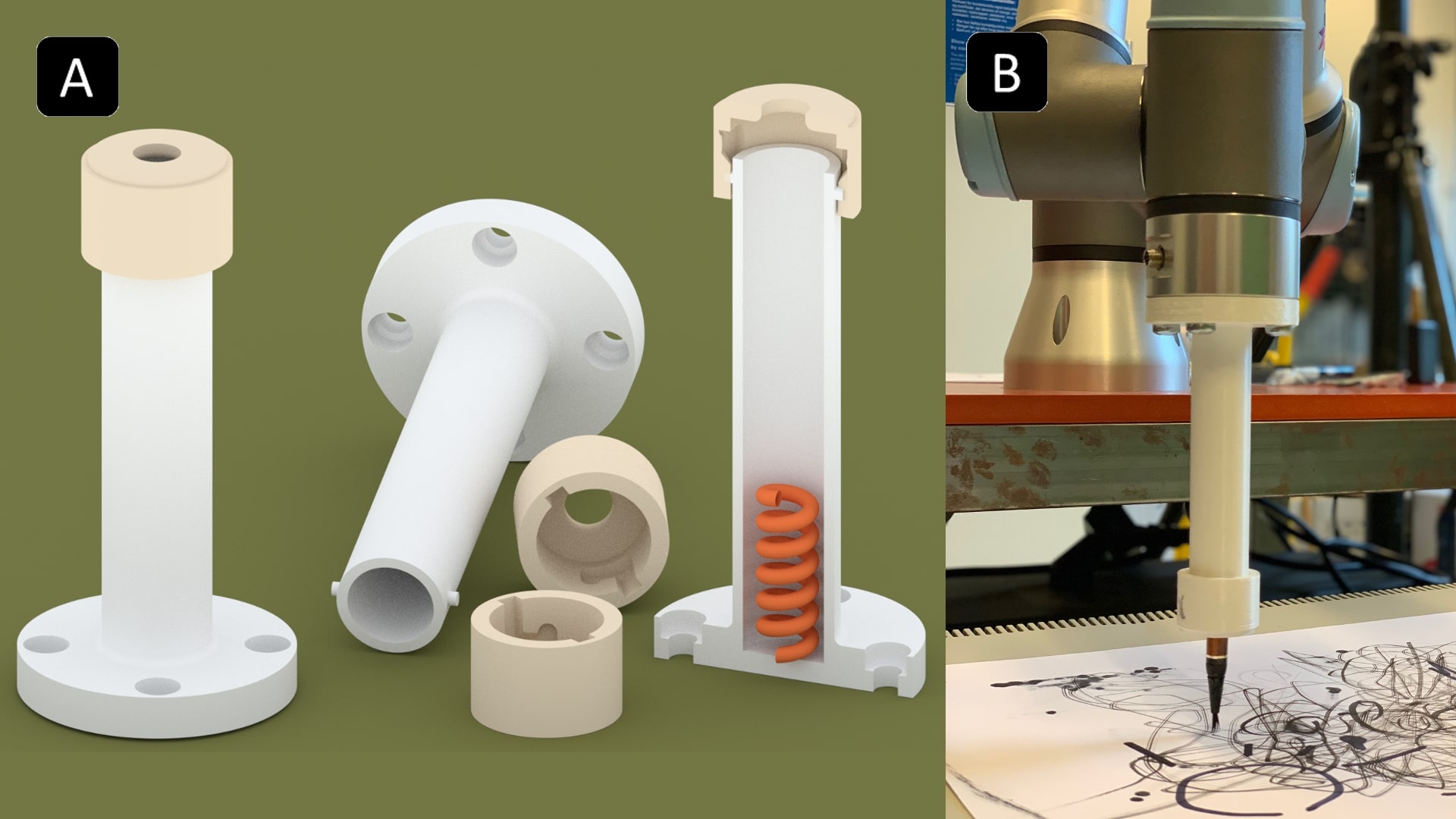}}
\caption{A, 3D model of the end effector. B, close look of the end effector mounted on the UR3 robot, loaded with an ink brush.}
\label{fig_endeffector}
\end{figure}

\subsection{End effector}
EMCAR repository includes a 3D-printable end effector specifically designed for drawing applications and attaches perfectly to the end of the robot arm with two M4 screws. The 3D model consists of a spring-loaded tube and a detachable cap that allows for quick tool changes. It is designed to hold standard ink markers securely, ensuring stable contact with the drawing surface during operation.
With this set-up, a variety of drawing tools have been tested, each offering distinct tactile and visual qualities. For example, markers produce sharp, consistent lines, more robotic, while brushes create more organic, fluid strokes with variable texture. Fig.\ref{fig_endeffector} shows the 3D model of the end effector, as well as an example loaded with an ink brush.

\subsection{User Interface}
\begin{figure}[tb]
\centerline{\includegraphics[width=\columnwidth]{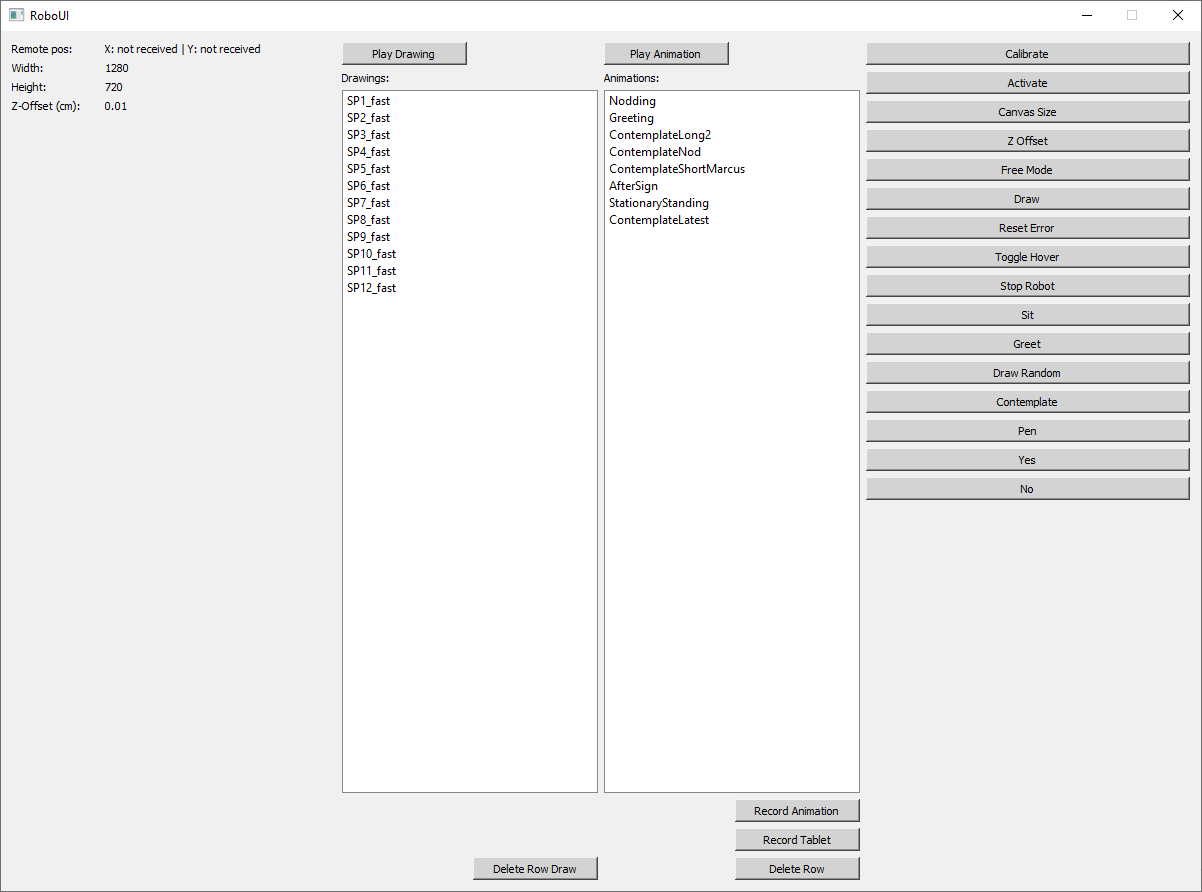}}
\caption{EMCAR's Graphic User Interface. In the center exists two columns with the drawings and animations saved. In the right side, the buttons to operate the software.}
\label{fig_GUI}
\end{figure}

The Graphical User Interface of EMCAR is a crucial component in making the software accessible, particularly for individuals without programming experience (see Fig.\ref{fig_GUI}). It provides an intuitive and user-friendly visual environment for controlling and configuring the robot's behavior.  
At the center of the interface are two frames: one for drawings and one for animations. These frames list all the sequences currently recorded. Below them, a set of buttons allows users to record new items or delete existing ones. At the top, a play button launches the selected item, streamlining the interaction and making the execution immediate and simple.

On the right of the GUI, a group of configuration buttons allows users to fine-tune the platform. Most of these are self-explanatory, but some deserve further explanation:

\begin{itemize}
    \item \textbf{Z Offset} adjusts the vertical position of the tool, allowing precise contact with the canvas, down to the millimeter. This is especially important when using different tools with different lengths, i.e. first using a long brush and after a sharpie. 
    \item \textbf{Canvas Size} sets the resolution of the drawing tablet, which affects spatial scaling during tele-operation.
    \item \textbf{Reset Error} is used to resume robot operation after an interruption, such as a collision or emergency stop.
\end{itemize}

Additionally, the last set of buttons can be customized to trigger specific drawings, animations, or even sequences of actions. These buttons pre-configured with default examples from the EMCAR repository and serve as useful templates for live demonstrations. With minor code modifications, users can create their own custom buttons by modifying the corresponding file. For details, refer to the documentation in the repository.

\subsection{Canvas calibration}
To ensure accurate spatial alignment between the robot and the drawing surface, EMCAR includes a guided canvas calibration process. The physical setup typically involves placing a leveled table in front of the robot, ideally mounted at a slightly lower height than the robot base, as presented in Fig.\ref{fig_workshop}. This ensures the robot reach in the drawing area and compensates for the length of the end effector. An A3 paper sheet can be fixed to the table using four 3D-printed corner brackets, which helps keep the canvas stationary throughout the drawing session.

Once the canvas is in place, calibrating begins by loading a drawing tool, such as a sharpie or ink brush, into the end effector. By clicking the \textbf{Calibrate} button in the GUI, the robot switches into \textbf{free mode} ( also known as "zero-G"), allowing it to be moved manually with minimal resistance.

The user is then prompted to touch the tip of the drawing tool on each corner of the canvas. As each point is registered, EMCAR captures the coordinates and uses them to define a coordinate frame that maps the tablet or drawing input to the robot's physical workspace.

After all four coordinates are set, the user confirms the calibration by clicking \textbf{OK} in the GUI. The system then saves this frame for subsequent drawing. This calibration persists after rebooting the system and only changes when a new calibration is performed. If needed, the \textbf{Z Offset} parameter can be adjusted to fine-tune the tool height, ensuring proper contact with the canvas without excessive pressure, or when the tool is changed to a longer one.

This physical, hands-on calibration approach supports EMCAR's goal: enabling intuitive interaction through movement and gesture. The process typically takes less than a minute and can be performed "on the fly" during workshops if the setup changes.

Once the table is placed and the corners glued, it is time to proceed with the calibration. First, the user must load the end effector with the tool, such as a Sharpie, and then click on \textbf{Calibrate} button. The robot will automatically switch to the free mode, also known as zero G, allowing it to move freely. gently grasp the end of the robot and follow the on-screen instructions; the GUI will prompt you to touch the tip of the Sharpie to each of the different corners of the canvas. Once this is complete, click OK, and the calibration is ready. If needed, the Z Offset button can be useful for fine-tuning the calibration.

\begin{figure}[tb]
\centerline{\includegraphics[width=\columnwidth]{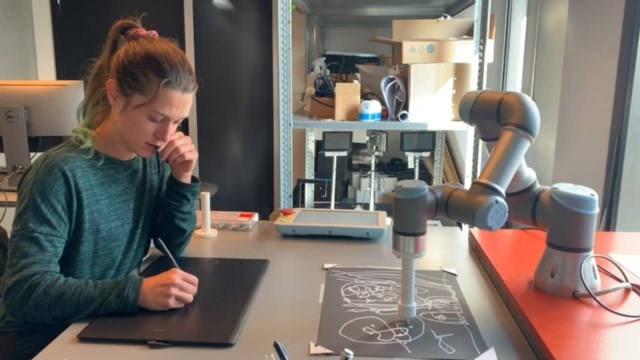}}
\caption{Artist in Residence Valeria Rizzo, during the development of EMCAR, testing the tele-operated drawing.}
\label{fig_Vali_painting}
\end{figure}

\subsection{Tablet and real time drawing}
The first of EMCAR's core features is its ability to translate real-time drawing gestures into robotics motion using a tablet. This interface enables users to tele-operate the robot intuitively by sketching directly on the calibrated canvas, with the robot mirroring these movements on the drawing surface (Fig.\ref{fig_Vali_painting}).

A Wacom drawing tablet is used as the input device. The tablet detects two states: when the stylus is hovering above the tablet and when it is in contact with it. This distinction is crucial. When the stylus hovers, the robot moves across the canvas, mirroring the stylus, but without making contact, positioning itself precisely for the next stroke. When the stylus touches the tablet, the robot lowers its tool to engage the drawing surface, allowing a continuous and controlled line to be drawn.

The low-latency introduced by the RTDE protocol becomes noticeable in this activity. The latency is due to the time the robot needs to receive the position, calculate the trajectory, and execute it. When the drawing is very detail and fast, the resulting layout at the robot side experiments a sort of low-pass filter, especially with drawing tools like sharpies or ball-pens.

In contrast, tools such as ink brushes, acrylic brushes, or even sponges offer their own material character, softening or obscuring this mechanical rigid trace. The resulting strokes appear more organic and expressive, often masking the fact that the drawing was done by a robot.

The process of recording a drawing is straightforward. In the GUI, the user first presses the \textbf{Free Mode} button. Now  the robot can be positioned so that the tool hovers a few centimeters above the canvas, pointing downward, as shown in Fig.~\ref{fig_endeffector} B.
Next, by pressing the \textbf{Draw} button a white window will appear, capturing the mouse pointer, which is now controlled by the tablet stylus. At this point, the stylus is already tele-operating the robot in real time. Finally, pressing the \textbf{Record Tablet} button starts saving the drawing gestures. Pressing \textbf{Record Tablet} again stops the recording. For convenience, the keyboard shortcut \textbf{R} can be used to turn the recording on and off.

\subsection{Tangible interaction and animation recording}
The second of EMCAR's core features is the ability to record animations as a set of robot movements directly through physical manipulation, a process that is intuitive and performative. The robot's free mode makes this possible, allowing for manual positioning of the different joints of the arm with minimal resistance, as a weightless limb.

To begin recording an animation, the user clicks the \textbf{Record Animation} button in the GUI. The robot automatically switches to free mode, allowing the user to grasp the arm and guide it through a sequence of motions. These movements are captured in real time and saved in a file as a set of joint angles that can later be replayed with the same speed, fluidity, and accuracy as while recording.

This mode of interaction allows the user to "puppeteer" the robot directly, treating it less like a machine to be programmed and more like a performative object to be animated. The technique resembles improvisational practices of puppetry, dance, and physical theater, where gestures are developed through embodied, physical exploration rather than programmatic coding. By physically shaping the robot's path, the user experiments the textile nature of the robot arm and applies their own rhythm and emotion.

\begin{figure}[tb]
\centerline{\includegraphics[width=\columnwidth, trim=0 500 0 150, clip]{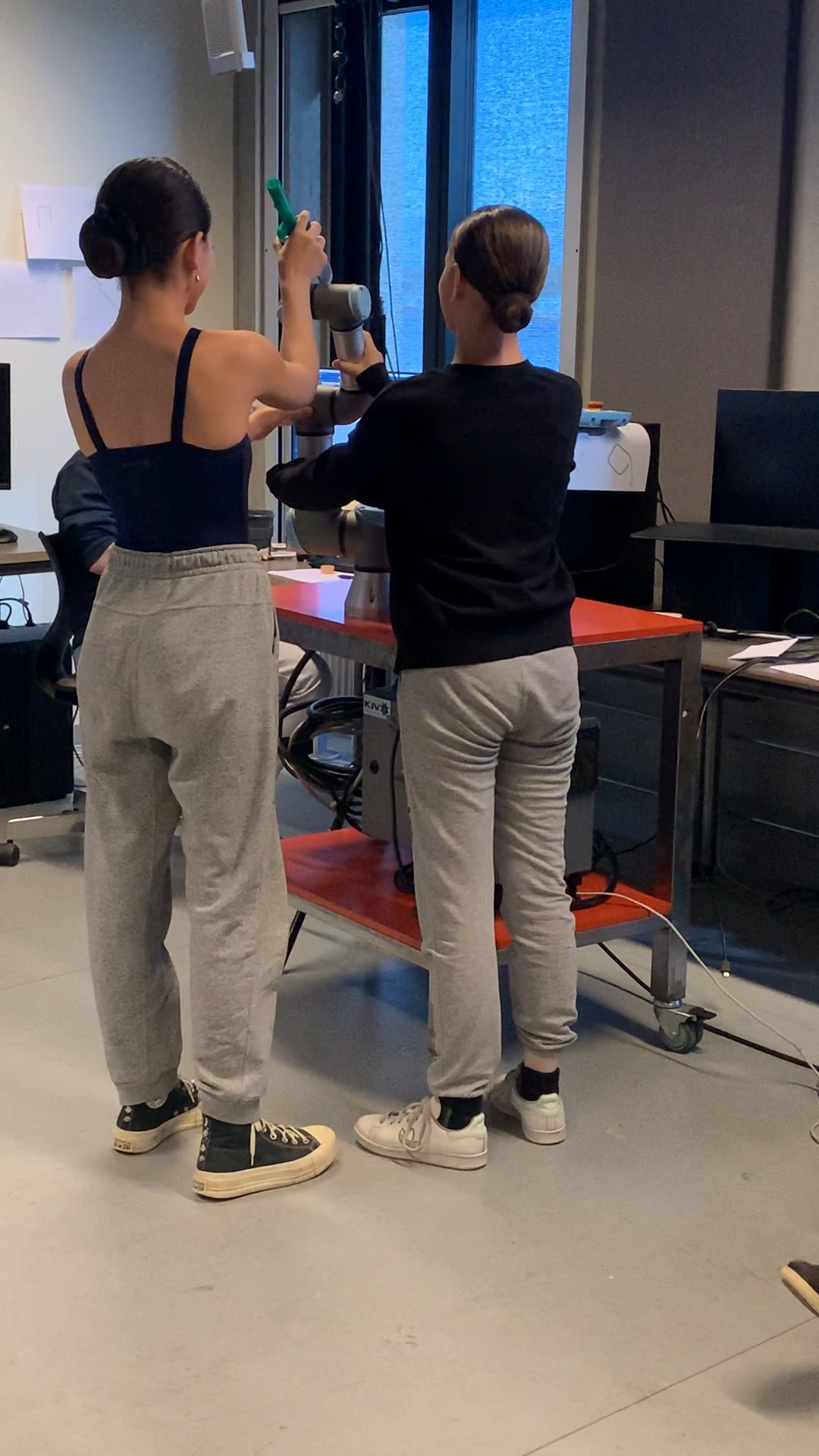}}
\caption{Ballet dancers recording the dancing movements of the robot using EMCAR while listening to the soundtrack}
\label{fig_Dancers_recording}
\end{figure}

\begin{figure}[tb]
\centerline{\includegraphics[width=\columnwidth]{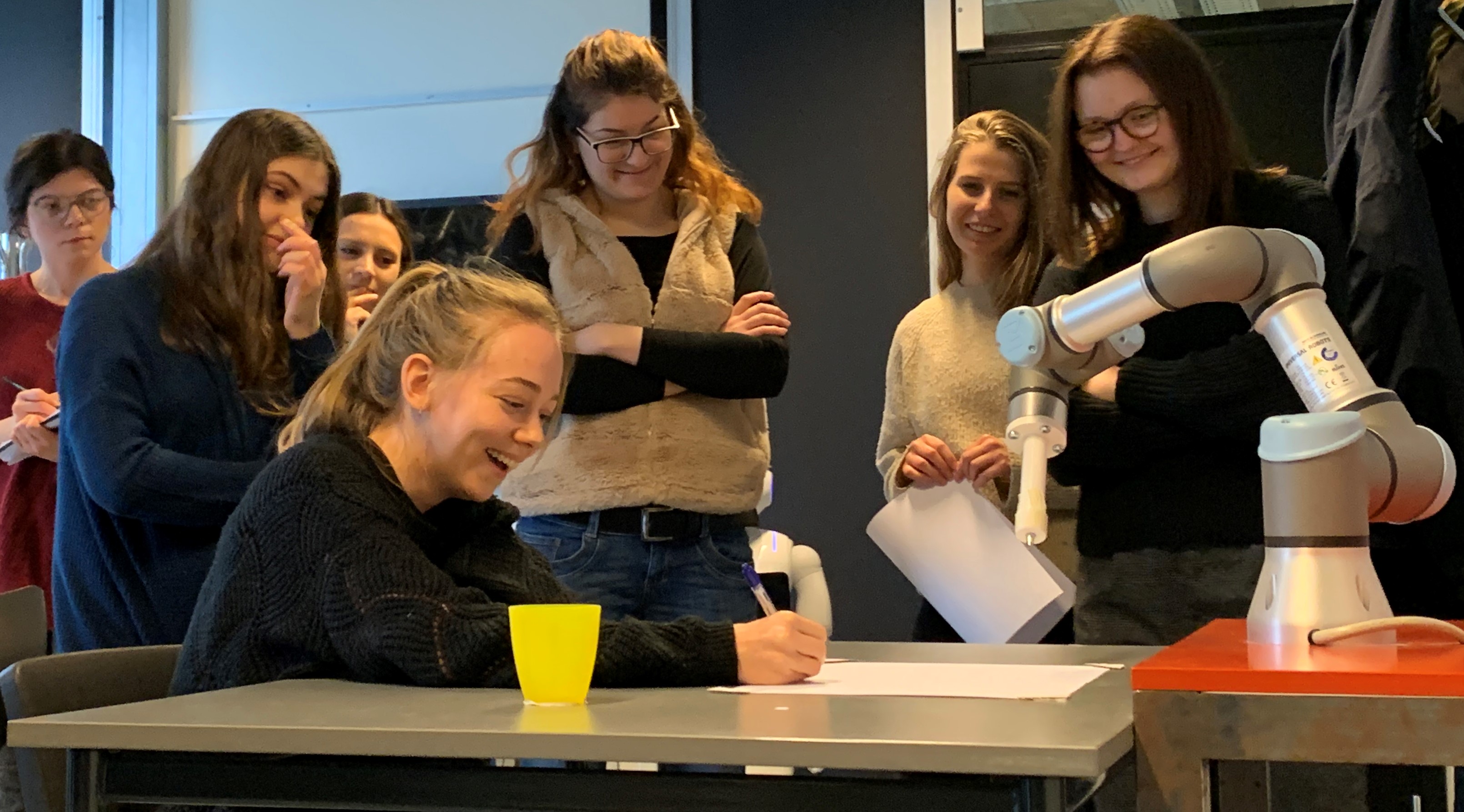}}
\caption{Drawing workshop using EMCAR, the robot is performing an animation, acting like observing the human action.}
\label{fig_workshop}
\end{figure}

The recorded animations can be played back immediately and iterated upon if needed, enabling rapid prototyping, experimentation, turn-tacking, and improvisation, essential qualities for the kind of creative context for which EMCAR was designed.
In Fig.\ref{fig_Dancers_recording} a pair of ballet dancers puppeteer the robot while listening to a soundtrack. The recorded animation is later played in sync with the music during a live ballet performance.

\subsection{Wizard of Oz}
Wizard of Oz is a well-established technique among researchers in Human-Robot Interaction \cite{10.5898/JHRI.1.1.Riek}. It consists of techniques for simulating autonomous behavior on the robot, having a human operator controlling it, often from behind the scenes, thereby creating the illusion that the robot possesses its own agency and decision-making capabilities. By pre-designing a range of interaction scenarios, researchers can study how humans respond to robot behaviors without needing a fully autonomous system. This approach is particularly useful during early design stages, when the primary focus is on the interaction design rather than the technical implementation.

The Wizard of Oz methodology is a fundamental principle in the design of EMCAR. All its core functions revolve around recording, saving, and playing animations or drawings. With minor code modifications, users can even chain a set of animations and/or drawings into a single button press, executing more complex, layered behaviors through simple, real-time cues.

\subsection*{}
For a quick visual overview on how to set and run EMCAR the Youtube video \textbf {"EMCAR: Embodied Controller for Animating Robots"} available at  \url{https://youtu.be/Lt0X3vqkKKw} explains it easily.

\section{Sample Projects}
Since its development, EMCAR has been used in a wide range of scenarios, including artistic performances, educational workshops, and open-to-the-public activities.
These projects have proved the versatility of the platform and its ability to engage with users, without regard to their technical background. We provide some brief examples that illustrate the diversity of EMCAR's applications.

\subsection {Drawing activities}
EMCAR has been used in collaborative drawing workshops with participants of all ages. These include open studio sessions for children at art museums and collaborative drawing workshops for university students. Drawing workshops were also conducted with adults living with mild to severe mental health disorders as part of a community Arts \& Health program, expanding the range of activities for socially assistive robots (SARs). Working closely with lay health workers and professional artists, we engaged a multi-phase, participatory design research program that included three pilot workshops followed by two interventions within an existing community mental health program. EMCAR was a key tool that enabled us to prioritize authentic stakeholder engagement and identify new use cases that include arts and cultural activities while addressing mental health. The outcomes of these these workshops are described in  \cite{ec6554510fc94f8aae45238901e9853b} \cite{d4df4c8a47c24676acbc0ead6c148268}.

\subsection {Puppetry}
\begin{figure}[tb]
\centerline{\includegraphics[width=\columnwidth]{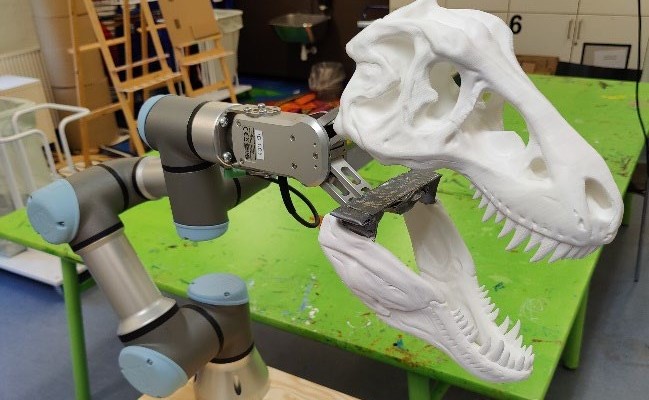}}
\caption{A 3D printed dinosaur skull attached to UR3, ready to be animated with EMCAR.}
\label{fig_dinosaur}
\end{figure}

In a creative robotics workshop held at a local community center, EMCAR was used to introduce the principles of robotics and how robots can be used as props for artistic creation. For this workshop, the drawing tool was replaced with a 3D-printed dinosaur skull Fig.\ref{fig_dinosaur}. This prop already helped attendees mimic the movement of a roaring Tyrannosaurus Rex, as a catalyst for improvisation and storytelling.

In another context, a group of students from the Bachelor's in Arts,  Art and Technology, picked EMCAR to puppeteer a UR5 robot for a theater performance, where the robot was playing one of the main characters. Using the Wizard of Oz approach to cue the animations that bring the robot to life.

\subsection {Dance Performances}
Working closely with the professional visual artist and choreographer Valeria Rizzo, EMCAR was first utilized in the development of a live dance performance titled If/Then. The performance is described in detail in \cite{alma9921065050005762}. A video of the full performance can be found at \cite{ifthen_2021}.

EMCAR was also used to co-develop a dance performance with a ballet school. Over a series of workshops, a group of young dancers learned the system and took turns adapting their own choreography to Camille Saint-Saën's Carnival of the Animals.  A recording of the performance can be accessed in Youtube \textbf{"Robot Ballet AAU /Aalborg Kulturskole"} available at
\url{https://youtu.be/NDTgYzXYWn4}

\begin{figure}[tb]
\centerline{\includegraphics[width=\columnwidth]{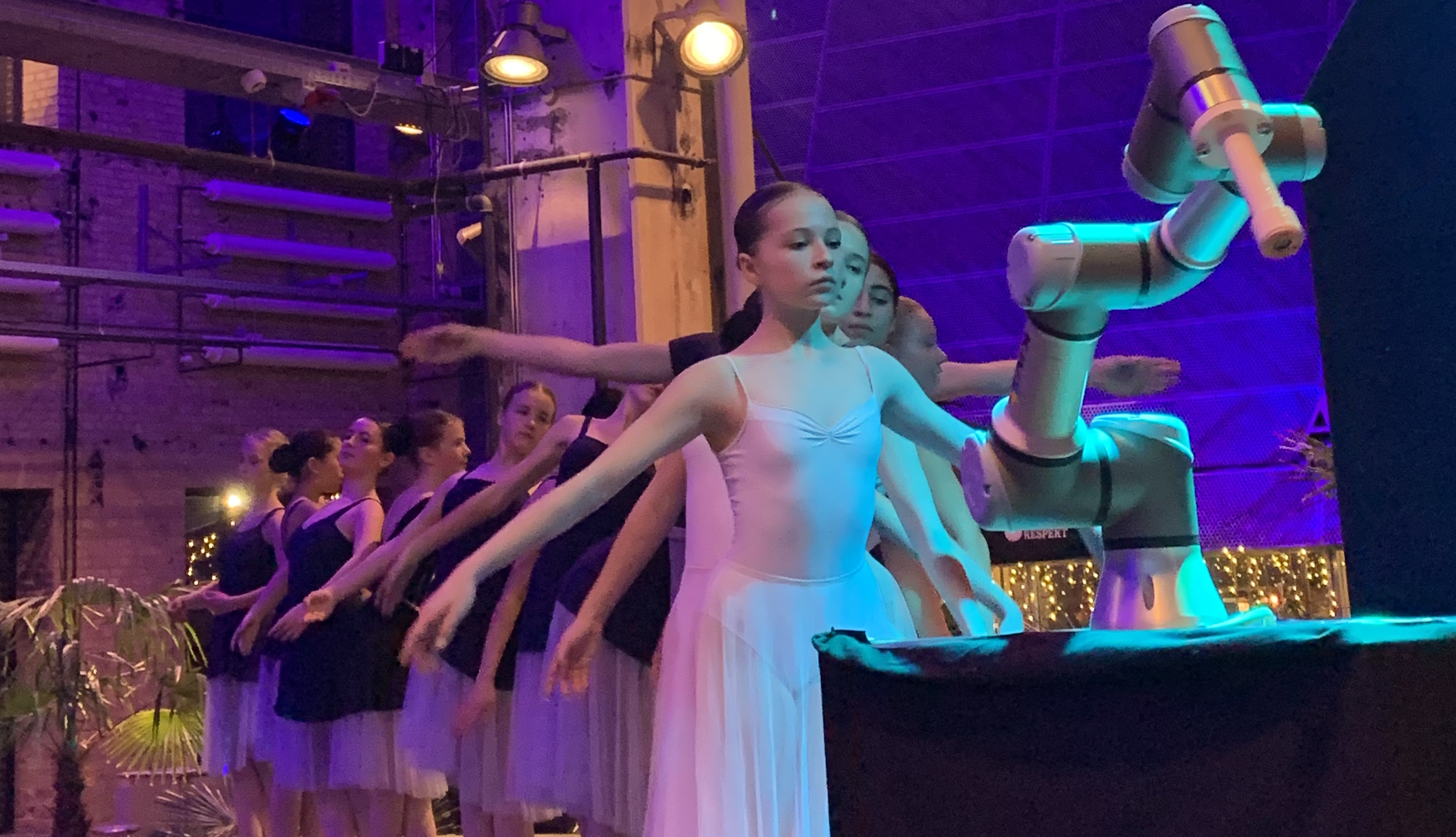}}
\caption{Ballet performance featured with a UR3 robot, controlled by EMCAR and animated by the ballet dancers.}
\label{fig_ballet}
\end{figure}

For each of the above cases, EMCAR  enabled artists and nontechnical users to explore intuitively the possibilities of what a robot can do through intuitive, hands-on interactions that allow people to interact with the system naturally. The advantage of working with systems that leverage embodied interaction is that individuals feel empowered to explore and also refine possible conceptual and material spaces with more ease. The tool makes important concepts in robotics - such as the need to calibrate - concrete without being overwhelming or feeling inaccessible.

\subsection {Teaching Platform}
EMCAR was utilised in a university course (titled Humans in the Loop: Robots, Automation, and Humanities Research), aimed at humanities and social science students with no previous programming experience to record, sketch, and test interaction scenarios with ease. The course explores how social science and humanities research frameworks can inform HRI research practice to contribute to more ethical, inclusive, and responsible robots and AI. Through research-based teaching and practice-based workshops, students learned to identify and apply practical quantitative and qualitative methods that bridge humanities research with relevant topics in engineering and computer science in their own mini-studies. EMCAR provided a hands-on experience that allowed non-robotics students to conceive and pilot new HRI research studies while working with an actual robot - a task that would have been impossible without a platform like EMCAR. We know from decades of literature and real world examples that understanding human factors like trust, expectations, and psychology are critical for designing good user experiences that will support widespread adoption of robots. EMCAR allowed students with no prior background in computer science or robotics to engage meaningfully in research studies and bring their own expertise and methodological frameworks to concrete HRI studies.    

\section{Conclusion}
 Literary theorist N. Kathrine Hayles insisted on the need to evaluate technical objects, especially digital tools, not only according to their function but as objects deeply embedded within larger social and technical processes. Hayles refers to "technical ensembles" - processes and practices through which fabrication comes about - wherein the toolmaker herself is embedded in both the practice and also in a society in which the knowledge of how to make tools is preserved, transmitted, and developed \cite{HowWeThink}. Computers and robots will continue to shape how we experience and make sense of the world and our place in it.  As interdisciplinary research becomes more commonplace, we hope that tools like EMCAR will contribute to valuable and meaningful collaborations between artists and roboticists, and help bridge the gaps to create more accessible pathways for creative and collaborative within HRI research and beyond.

\bibliographystyle{IEEEtran}
\bibliography{bibliography.bib}





\end{document}